\documentclass[sort,compress,11pt]{elsarticle}

\usepackage{amssymb}
\usepackage{amsthm}
\usepackage{amsmath}
\usepackage{mathtools}
\usepackage{mathrsfs}
\usepackage[ruled, lined, linesnumbered, longend]{algorithm2e}
\SetKw{init}{Initialization:}
\let\oldnl\nl
\newcommand{\nonl}{\renewcommand{\nl}{\let\nl\oldnl}}
\usepackage{array}
\usepackage{multirow}
\usepackage{listings}
\usepackage{tabu}
\usepackage{enumerate}
\usepackage{lineno}
\usepackage{fullpage}
\usepackage{xcolor}
\usepackage[colorinlistoftodos]{todonotes}
\usepackage{bbm}
\usepackage[colorlinks=true]{hyperref}
\usepackage{url}
\usepackage{textcomp}
\usepackage{gensymb}
\usepackage{soul}
\usepackage{graphicx}
\usepackage{subfig}

\usepackage{fourier} 
\usepackage{array}
\usepackage{makecell}
\usepackage[export]{adjustbox}
\usepackage{float}
\usepackage{caption}
\usetikzlibrary{shapes}
\biboptions{numbers,comma,round,square}
\usepackage{arydshln}
\usepackage[para,online,flushleft]{threeparttable}

\usepackage{float}
\usepackage{booktabs}
\usepackage{tabularx}
\usepackage{array}

\newcolumntype{L}{>{\raggedright\arraybackslash}X}
\newcolumntype{P}[1]{>{\raggedright\arraybackslash}p{#1}}

\usepackage{orcidlink}

\definecolor{myred}{RGB}{214,39,40}
\definecolor{mygray}{RGB}{176,176,176}
\definecolor{myorange}{RGB}{255,127,14}
\definecolor{mygreen}{RGB}{44,160,44}
\definecolor{mylightgray}{RGB}{204,204,204}
\definecolor{mypurple}{RGB}{148,103,189}
\definecolor{mybrown}{RGB}{140,86,75}
\definecolor{steelblue}{RGB}{31,119,180}
\definecolor{intraray}{RGB}{127,193,219}
\definecolor{hemitrichs}{RGB}{26,74,93}

\theoremstyle{definition}

\theoremstyle{remark}

\graphicspath{ {./figs/} }

\journal{Elsevier}
\DontPrintSemicolon
\definecolor{orcidlogocol}{HTML}{A6CE39}
\begin{document}
\makeatletter
\def\ps@pprintTitle{%
  \let\@oddhead\@empty
  \let\@evenhead\@empty
  \let\@oddfoot\@empty
  \let\@evenfoot\@oddfoot
}
\makeatother
\begin{frontmatter}

\title{PiDDM: Physics-Informed Differentiable Degradation Modeling for Lithium-Ion Battery State-of-Health Prediction}

\author[ndCBE]{Zeping Chen}
\author[UTD]{Ruda Jian}
\author[UTD]{Sachin Sigdel}
\author[UTD]{Guoping Xiong}
\author[ndAME,cornell]{Jian-Xun Wang}
\author[ndCBE,ndAME]{Tengfei Luo\corref{corjw}}

\address[ndCBE]{Department of Chemical and Biomolecular Engineering, University of Notre Dame, Notre Dame, IN, USA}
\address[UTD]{Department of Mechanical Engineering, The University of Texas at Dallas, Richardson, TX, USA}
\address[ndAME]{Department of Aerospace and Mechanical Engineering, University of Notre Dame, Notre Dame, IN, USA}
\address[cornell]{Sibley School of Mechanical and Aerospace Engineering, Cornell University, Ithaca, NY, United States of America}

\cortext[corjw]{Corresponding author: tluo@nd.edu}

\begin{abstract}
Accurate prediction of lithium-ion battery state-of-health (SOH) is essential for the reliable operation of modern energy storage systems. However, purely data-driven machine learning models often generalize poorly across different cycling protocols and may produce physically implausible predictions during long-term extrapolation. To address these limitations, we developed a physics-informed differentiable degradation modeling framework (PiDDM) for lithium-ion battery SOH degradation prediction. The proposed framework incorporates empirical Arrhenius degradation kinetics associated with solid electrolyte interphase (SEI) growth and loss of lithium inventory (LLI) into the neural network training objective. By embedding this degradation regularization into the loss function, PiDDM encourages physically consistent monotonic capacity fade under diverse operating conditions. We evaluate the model on a public dataset containing 55 batteries cycled under six distinct operating protocols. PiDDM achieves the lowest average prediction error compared to other existing models, outperforming both a standard multilayer perceptron (MLP) and a baseline physics-informed neural network (PINN), with a substantially reduced mean squared error. In extrapolation tests, the models are trained on only the first 90\% of each battery’s cycle life and evaluated on the remaining unseen 10\%. Under this setting, PiDDM accurately captures the accelerated end-of-life degradation while avoiding the non-physical capacity regeneration observed in the baseline models. These results demonstrate that incorporating degradation physics into neural network training improves both predictive accuracy and physical consistency, providing a promising framework for practical battery health monitoring.
\end{abstract}
\end{frontmatter}

\section{Introduction}

Lithium-ion batteries have become the dominant rechargeable energy storage technology worldwide, supporting electrified transportation, consumer electronics, and an increasing share of stationary storage for renewable energy integration~\cite{Tarascon2001,Armand2008,Dunn2011,Goodenough2013,Nitta2015}. Their importance continues to grow as electric vehicle adoption accelerates. In 2023, nearly 14 million new electric cars were registered globally, representing a 35\% increase relative to 2022~\cite{IEA_GEVO2024_cars}. This growth is driving a rapid rise in battery demand. The International Energy Agency estimated that demand for electric-vehicle batteries exceeded 750\,GWh in 2023, a 40\% increase compared with 2022~\cite{IEA_GEVO2024_batteries}. These trends underscore the central role of lithium-ion batteries in modern energy systems. However, their performance and service life remain constrained by aging and degradation. During storage and cycling, capacity decreases and internal resistance increases, reducing usable energy and power while potentially increasing safety risks. Understanding how degradation evolves under different operating protocols is, therefore, essential for improving battery utilization, enabling reliable health management, and designing charging and discharging strategies that extend service life~\cite{Vetter2005,Barre2013,Birkl2017,Edge2021,OKane2022}.

Models are available for predicting battery life but face persistent challenges in practical deployment. For example, physics-based models, particularly porous electrode models, provide mechanistic descriptions of coupled transport and electrochemical processes and can help interpret degradation behavior in terms of underlying physical mechanisms~\cite{Newman1975,Doyle1993,Fuller1994,ChenAEM2022_porous_review,OKane2022}. However, their application in real-world settings often requires extensive parameterization and calibration, and many of the required parameters are difficult to identify from routine measurements~\cite{ChenAEM2022_porous_review,Yu2025_state_estimation_review,Sulzer2021PyBaMM}. As a result, physics-based approaches are frequently combined with estimation or data-driven components to compensate for limited observability, highlighting the difficulty of relying on physics alone for robust battery state tracking in practice~\cite{Plett2004EKF,Yu2025_state_estimation_review}. In parallel, data-driven methods have shown considerable promise for degradation forecasting, but they often struggle to generalize when operating conditions differ from those represented in the training data, particularly under accelerated aging or protocol shifts~\cite{Severson2019_MIT,Attia2020,Richardson2017,Roman2021,Sadler2025}. These limitations motivate the development of methods that preserve physical grounding while improving robustness across diverse operating conditions.

Deep learning has advanced rapidly across a wide range of scientific and engineering disciplines, including manufacturing~\cite{Akhare2024}, computational mechanics~\cite{CHEN_composite}, and battery modeling~\cite{Severson2019_MIT,Attia2020,Richardson2017,Roman2021,Wang_NC,Ma_HUST,Zhu2022_TJU}. In the battery field, the increasing availability of public datasets~\cite{Severson2019_MIT,Attia2020,Ma_HUST,Zhu2022_TJU,Wang_NC} and the ability of neural networks to learn directly from voltage, current, and time-series measurements have made machine learning an important direction for state-of-health (SOH) prediction. Many studies have reported strong predictive accuracy under controlled conditions and within specific cycling protocols~\cite{Severson2019_MIT,Attia2020,Richardson2017,Roman2021,Wang_NC,Ma_HUST,Zhu2022_TJU}. However, performance often degrades when models are evaluated outside the training distribution, especially when both charging and discharging protocols vary. This limitation is critical for real-world applications, where operating conditions differ substantially across users, climates, and use cases. A key challenge is therefore to develop learning-based health models that remain accurate, stable, and physically credible across a broad range of cycling protocols without extensive retraining for each new condition.

To address this challenge, we developed Physics-informed Differentiable degradation modeling (PiDDM) for battery SOH prediction. Unlike conventional physics-informed neural networks (PINNs) that impose generic residual penalties~\cite{Raissi2019,Karniadakis2021}, PiDDM embeds battery-aging kinetics as a degradation-direction constraint that directly regularizes the temporal evolution of SOH. In particular, the model accounts for degradation associated with solid electrolyte interphase growth and loss of lithium inventory~\cite{Peled2017,PinsonBazant2013,Edge2021,OKane2022}, thereby encouraging physically consistent capacity fade under diverse operating conditions. We evaluate PiDDM on a public dataset of 55 batteries~\cite{Wang_NC} cycled under six distinct operating protocols, ranging from constant-current conditions to highly dynamic load profiles. Compared with a standard multilayer perceptron (MLP) and a baseline PINN, PiDDM achieves lower average prediction error and improved extrapolation performance. In particular, when the models are trained only on the first 90\% of each battery's cycle life and evaluated on the remaining unseen 10\%, PiDDM captures the accelerated end-of-life degradation more accurately while avoiding the non-physical capacity regeneration observed in the baseline models. These results demonstrate the value of incorporating degradation physics into neural network training for more robust and physically consistent battery health prediction.


\section{Methodology}

\subsection{Battery degradation mechanisms and physics-informed degradation-rate }

Lithium-ion battery aging arises from coupled electrochemical, transport, and interfacial side reactions that progressively reduce usable capacity and increase internal resistance~\cite{Vetter2005,Barre2013,Birkl2017,Edge2021,OKane2022}. A complete mechanistic description of cell operation can be constructed using porous-electrode models, particularly the Doyle--Fuller--Newman framework, which resolves electrolyte transport, interfacial reaction kinetics, and solid-phase lithium diffusion at the electrode scale~\cite{Newman1975,Doyle1993,Fuller1994,ChenAEM2022_porous_review}. However, full physics-based degradation models require extensive parameterization, including solid-phase diffusivities, electrolyte transport properties, reaction-rate constants, active surface areas, and degradation-specific kinetic parameters. Many of these quantities are difficult to identify uniquely from routine cycling measurements such as voltage, current, temperature, and capacity. This limitation motivates the use of semi-empirical degradation representations that preserve physically meaningful aging trends while remaining compatible with data-driven learning~\cite{OKane2022,Yu2025_state_estimation_review}.

At the cell level, lithium-ion battery degradation is commonly interpreted through measurable degradation modes, including loss of lithium inventory (LLI), loss of active material at the positive electrode, loss of active material at the negative electrode, and impedance rise~\cite{Vetter2005,Barre2013,Birkl2017,Edge2021,OKane2022}. Among these mechanisms, LLI associated with continuous solid electrolyte interphase (SEI) growth on the graphite negative electrode is widely recognized as a major contributor to capacity fade under conventional graphite-based lithium-ion operating conditions~\cite{Peled1979,ChristensenNewman2005,Peled2017,PinsonBazant2013,Ramadass2003CapacityFade,Keil2016}. The SEI forms because the graphite anode operates at potentials where common organic carbonate electrolytes are thermodynamically unstable. During storage and cycling, electrolyte-reduction products accumulate on the anode surface and form a passivating interphase. Although this interphase is necessary for stable cell operation, its continued growth irreversibly consumes cyclable lithium, increases interfacial resistance, and reduces extractable cell capacity~\cite{Ploehn2004,Ramadass2004,Single2018}.

The rate of these parasitic degradation reactions is strongly affected by temperature. This thermal acceleration is commonly represented using an Arrhenius-type relation,
\[
r = k_0 \exp\!\left(-\frac{E_a}{RT}\right),
\]
where $r$ is the reaction-rate term, $k_0$ is the pre-exponential factor, $E_a$ is the apparent activation energy, $R$ is the universal gas constant, and $T$ is the absolute temperature. In lithium-ion battery aging models, Arrhenius-type relations are widely used to describe the increase in degradation rate at elevated temperature~\cite{Bloom2001,Wang2011,Schmalstieg2014,Schmalstieg2014HolisticAging,Keil2016,Leng2015}. In addition to temperature, capacity fade depends on usage severity, including depth of discharge ($DoD$), accumulated ampere-hour throughput ($Ah$), C-rate, and cycling-profile variability~\cite{Wang2011,Schmalstieg2014HolisticAging,Lin2013CapacityFade}. Semi-empirical cycle-life models commonly represent these dependencies using power-law terms in time or charge throughput combined with Arrhenius temperature acceleration~\cite{Wang2011,Schmalstieg2014,Schmalstieg2014HolisticAging,Keil2016}.

The degradation relation used in this work is therefore designed to act as a physics-informed learning prior rather than a fully prescribed electrochemical model. It must satisfy three requirements. First, it should preserve the physically expected direction of capacity evolution, namely monotonic capacity fade rather than non-physical capacity regeneration. Second, it should encode the main operating-condition dependencies known to accelerate aging, including temperature, time, depth of discharge, and accumulated charge throughput. Third, it should remain differentiable so that it can be embedded directly into the neural-network training objective and optimized through gradient-based learning. A purely data-driven model can learn local correlations from cycling data, but it may produce physically implausible trajectories under extrapolation or protocol shift~\cite{Severson2019_MIT,Attia2020,Richardson2017,Roman2021,Wang_NC}. In contrast, the semi-empirical degradation-rate prior used here provides an intermediate formulation: it does not resolve all electrode-scale degradation pathways, but it constrains the learned SOH trajectory using physically meaningful degradation structure.

The Arrhenius--power-law form in  Eq.~\eqref{eq:arrhenius_degradation} was selected as a reduced-order kinetic prior because its components have physically interpretable origins while remaining compatible with cycling-level measurements. For SEI-related degradation, the Arrhenius temperature dependence is motivated by transition-state and reaction-rate theory, according to which thermally activated interfacial reactions exhibit an exponential dependence on inverse absolute temperature. The time-dependent power-law term accommodates the sublinear behavior commonly associated with transport-limited SEI growth, while the power-law dependencies on depth of discharge and accumulated charge throughput follow established semi-empirical cycle-life models used to represent cycling severity and cumulative exposure. This formulation is not claimed to be a unique or complete description of battery aging. It does not explicitly resolve interactions among degradation mechanisms, including lithium plating, loss of active material, impedance growth, or changes in the dominant mechanism under extreme temperatures and C-rates. Therefore, the learned coefficients should be interpreted as effective, data-calibrated parameters within the operating conditions represented by the training data. Based on these considerations, the capacity-fade rate is represented as the sum of an SEI-associated time-dependent contribution and an LLI-associated cycling-severity contribution:
\begin{equation}
\frac{dQ}{dt}
=
-k_{SEI}\exp\!\left(-\frac{E_{SEI}}{RT}\right)t^{\alpha}
-k_{LLI}\exp\!\left(-\frac{E_{LLI}}{RT}\right)DoD^{b_1}Ah^{b_2}.
\label{eq:arrhenius_degradation}
\end{equation}
Here, $Q$ denotes battery capacity, $t$ is time, $T$ is absolute temperature, $R$ is the universal gas constant, $DoD$ is the depth of discharge, and $Ah$ is the accumulated ampere-hour throughput. The coefficients $k_{SEI}$ and $k_{LLI}$ represent effective degradation-rate factors associated with SEI-dominated aging and LLI-related cycling degradation, respectively. The apparent activation energies $E_{SEI}$ and $E_{LLI}$ describe the thermal sensitivity of the two degradation contributions, while $\alpha$, $b_1$, and $b_2$ describe the empirical sensitivity of degradation to time, depth of discharge, and accumulated charge throughput.

In the proposed PiDDM framework, Eq.~\eqref{eq:arrhenius_degradation} is used as a differentiable degradation-rate prior during model training. The neural network is not required to operate as a complete physics-based battery model. Instead, it learns effective degradation parameters and latent operating-state dependencies from cycling features, while the physics-informed term regularizes the temporal evolution of predicted capacity. The negative sign in Eq.~\eqref{eq:arrhenius_degradation} enforces the physically expected direction of capacity loss, while the Arrhenius and power-law terms encode thermal acceleration and cycling-severity effects. This formulation therefore integrates degradation physics with machine learning by using physical knowledge as an inductive bias rather than as a rigid forward model~\cite{Raissi2019,Karniadakis2021}.

Equation~\eqref{eq:arrhenius_degradation} is not intended to resolve every electrode-scale degradation pathway explicitly. Mechanisms such as loss of active material at the positive electrode, loss of active material at the negative electrode, lithium plating, transition-metal dissolution, cathode structural degradation, and electrolyte oxidation are not separately parameterized in the present formulation~\cite{Edge2021,OKane2022,Lin2013CapacityFade}. Instead, their effects are treated as unresolved contributions that may be absorbed into the learned degradation coefficients when their signatures appear in the training data. This formulation provides a compact physics-informed regularization constraint: it preserves the monotonic and thermally activated character of lithium-ion battery aging while remaining compatible with cycling-level measurements used for SOH prediction.

\subsection{Battery degradation modeling}

\begin{figure}[!ht]
\centering
\includegraphics[width=\textwidth]{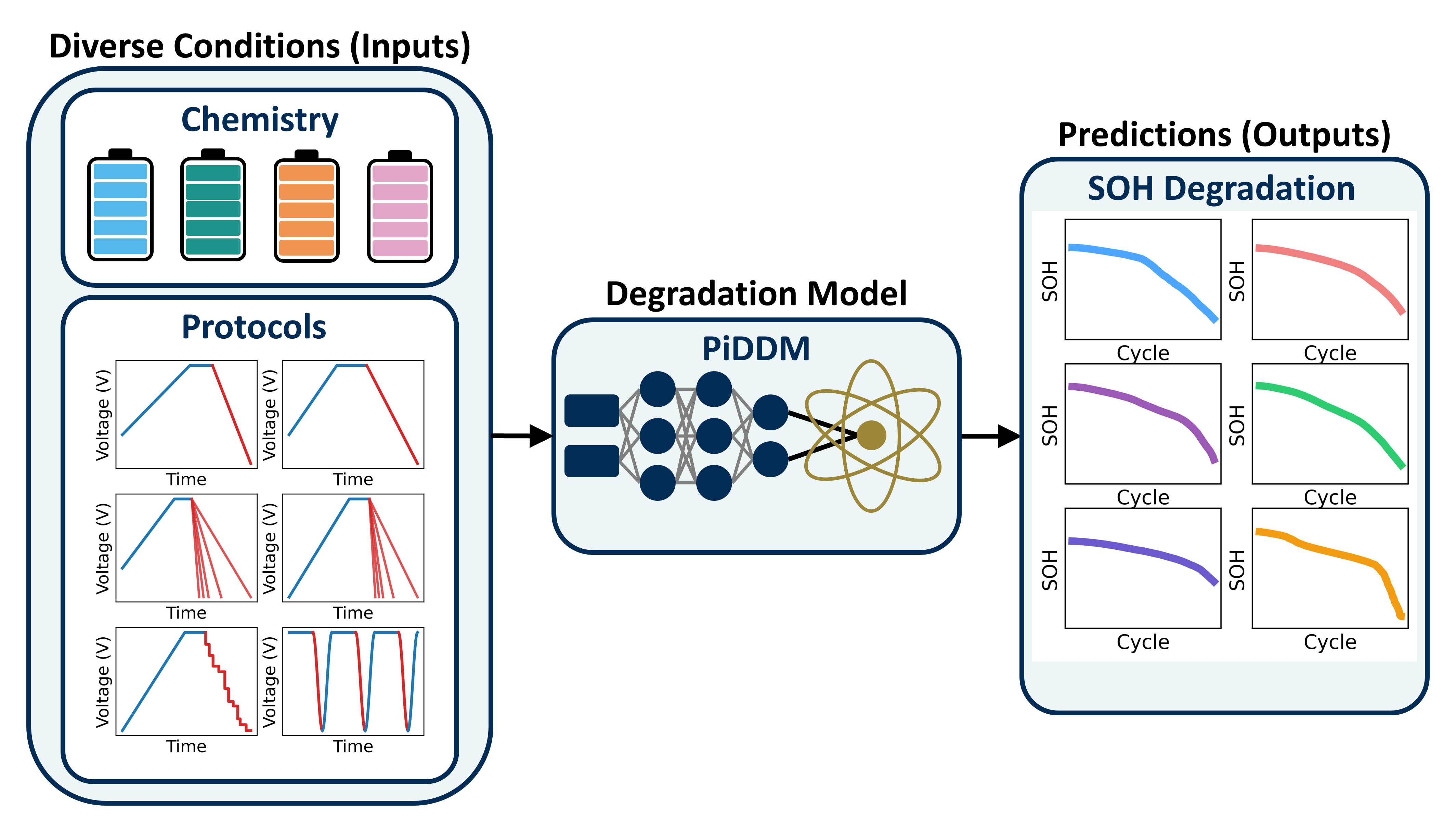}
\caption{\textbf{Overview of the Physics-informed Differentiable degradation modeling (PiDDM) framework for battery state-of-health (SOH) prediction.} Charging and discharging protocol information is processed by the PiDDM, which combines data-driven learning with empirical degradation kinetics to predict physically consistent SOH trajectories over the battery lifetime.}
\label{fig:overview}
\end{figure}

To connect the degradation mechanisms described above with observable battery behavior, we develop a surrogate modeling framework that integrates empirical degradation kinetics with data-driven learning. Figure~\ref{fig:overview} presents an overview of the proposed framework. The proposed framework is designed to model battery degradation across diverse operating conditions. Instead of treating the relationship between operational inputs and state of health (SOH) as a purely black-box mapping, PiDDM embeds empirical aging information into the learning process. In this way, the model links measurable cycling behavior with degradation-governing factors, allowing it to better capture SOH evolution under different charging and discharging protocols. This integration improves the physical consistency of the predicted degradation trajectories and enhances the model's ability to generalize to previously unseen regions.

\subsection{PiDDM Framework}
\label{sec:piddm_framework}

\begin{figure}[!ht]
\centering
\includegraphics[width=\textwidth]{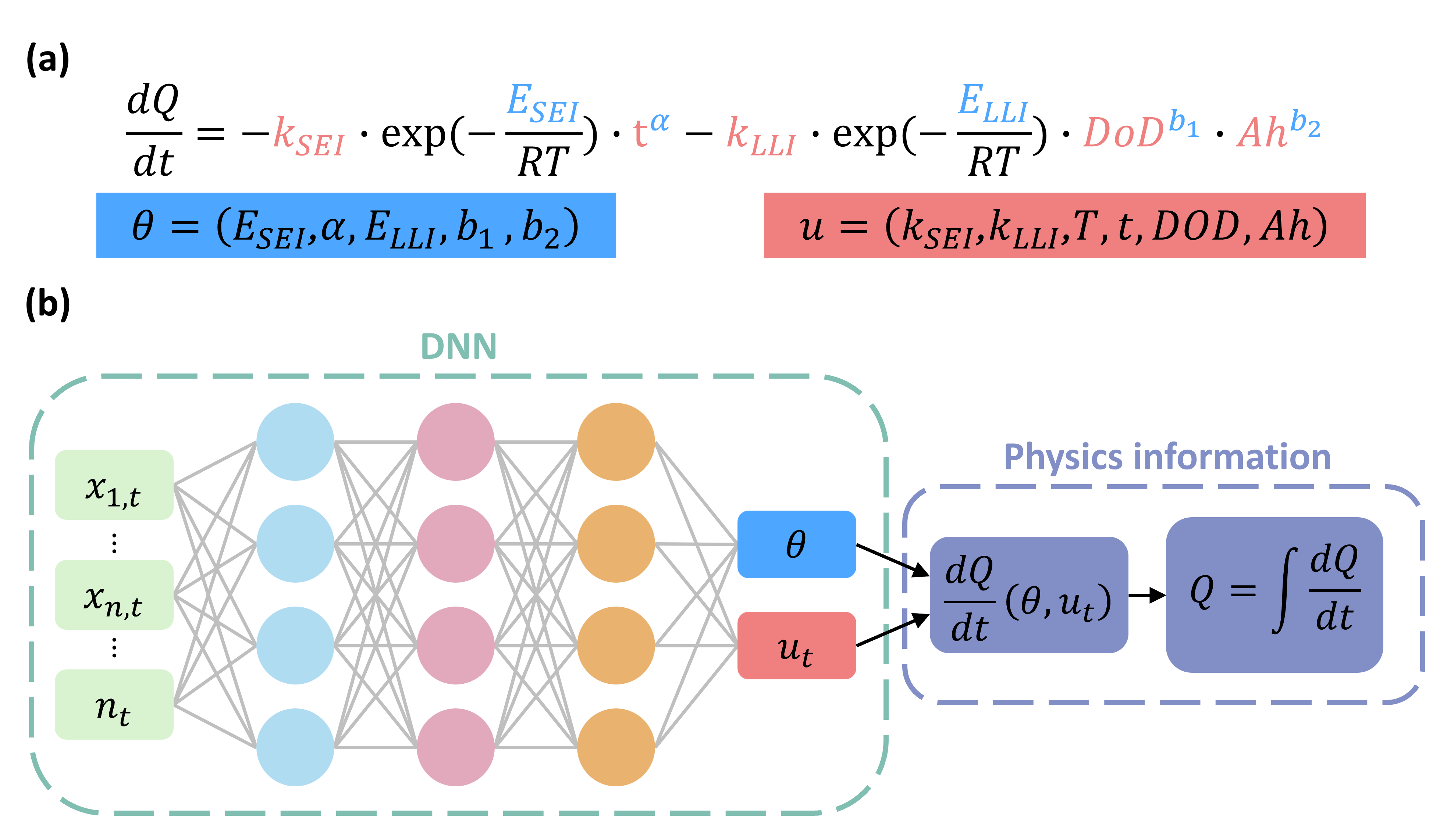}
\caption{\textbf{Physics-informed Differentiable degradation modeling (PiDDM).} (a) Empirical capacity-fade relation containing SEI-associated and LLI-associated degradation contributions. (b) The neural network maps the 16-dimensional charging-feature vector $\mathbf{x}_n$ to five effective, cycle-dependent kinetic parameters $\widehat{\boldsymbol{\theta}}_n$. These parameters are combined with the global trainable pre-exponential factors $\boldsymbol{\kappa}$ and the known operating variables $\mathbf{z}_n$ in the differentiable degradation module. Temporal integration of the resulting degradation rate produces the predicted capacity trajectory.}
\label{fig:model_framework}
\end{figure}

Purely data-driven neural networks can capture complex degradation patterns within the training distribution, but they often become unreliable when extrapolated beyond the observed data. To improve extrapolation performance and physical consistency, PiDDM incorporates the Arrhenius--power-law degradation relation introduced in Eq.~\eqref{eq:arrhenius_degradation} as a differentiable physics-based constraint during training. The overall model structure is illustrated in Figure~\ref{fig:model_framework}.

For cycle \(n\), let \(\mathbf{x}_n\in\mathbb{R}^{16}\) denote the normalized charging-feature vector provided to PiDDM. The neural network, with trainable weights and biases \(\boldsymbol{\phi}\), produces the five-dimensional output
\[
\widehat{\boldsymbol{\theta}}_n
=
f_{\boldsymbol{\phi}}(\mathbf{x}_n)
=
\left(
\widehat{E}_{\mathrm{SEI},n},
\widehat{\alpha}_n,
\widehat{E}_{\mathrm{LLI},n},
\widehat{b}_{1,n},
\widehat{b}_{2,n}
\right).
\]
These quantities are effective, cycle-dependent kinetic parameters inferred from the charging features. They are distinct from the neural-network weights and biases \(\boldsymbol{\phi}\) and should not be interpreted as independently measured electrochemical parameters. Their values describe the degradation behavior represented by the training data and may therefore depend on the operating regime over which the model is trained.

The two pre-exponential degradation-rate factors are denoted by
\[
\boldsymbol{\kappa}
=
\left(
k_{\mathrm{SEI}},
k_{\mathrm{LLI}}
\right).
\]
Both \(k_{\mathrm{SEI}}\) and \(k_{\mathrm{LLI}}\) are global trainable scalar parameters. They are initialized before training and optimized jointly with \(\boldsymbol{\phi}\) through the PiDDM training objective. They are not prescribed from literature values or independently predicted by the neural network at each cycle; instead, each coefficient is shared across all batteries and cycles within a given trained model.

The remaining quantities entering Eq.~\eqref{eq:arrhenius_degradation} are known operating variables, which are grouped as
\[
\mathbf{z}_n
=
\left(
T_n,t_n,\mathrm{DoD}_n,Ah_n
\right).
\]
Here, \(T_n\) and \(\mathrm{DoD}_n\) are obtained from the operating protocol, whereas \(t_n\) and \(Ah_n\) denote the elapsed time and accumulated ampere-hour throughput calculated from the cycling record. The universal gas constant \(R\) is fixed. None of the quantities in \(\mathbf{z}_n\) is predicted by the neural network.

As shown in Figure~\ref{fig:model_framework}(b), the input vector \(\mathbf{x}_n\) consists of 16 statistical features extracted from the regular charging segment. The features are derived from a short, fixed window of the constant-current (CC) charging stage and the subsequent constant-voltage (CV) stage, including the voltage response near the cut-off threshold and the current decay during the CV period~\cite{Severson2019_MIT,Attia2020,Ma_HUST,Zhu2022_TJU,Wang_NC}. They include the mean, standard deviation, kurtosis, skewness, charging time, accumulated charge, curve slope, and entropy-related descriptors of the measured charging signals~\cite{Severson2019_MIT,Attia2020,Ma_HUST,Zhu2022_TJU,Wang_NC}. Because these features are constructed only from routinely available charging measurements, they do not directly include discharge-capacity information or the target SOH trajectory. This reduces target leakage and facilitates application across batteries operated under previously unseen discharge protocols~\cite{Severson2019_MIT,Attia2020,Wang_NC}. In practical applications, the required features could be collected through standardized diagnostic charging measurements. The baseline models, neural-network architectures, training hyperparameters, and evaluation protocols are described in Section~\ref{sec:baseline_training_evaluation}.

The model is trained and evaluated using the public battery dataset reported by Wang et al.~\cite{Wang_NC}. The dataset contains 55 lithium-nickel-cobalt-manganese-oxide (NCM) batteries cycled to failure under six distinct charge and discharge protocols. These protocols include constant-current conditions, resistance-based aging conditions, random-walk loading, and geosynchronous earth orbit (GEO) satellite mission profiles, providing a challenging benchmark for evaluating model generalization across diverse operating conditions~\cite{Severson2019_MIT,Attia2020,Wang_NC}.

Within the differentiable degradation module, the network-predicted kinetic parameters \(\widehat{\boldsymbol{\theta}}_n\), the global trainable coefficients \(\boldsymbol{\kappa}\), and the known operating variables \(\mathbf{z}_n\) are combined to evaluate the capacity-fade rate:
\[
\widehat{\dot{Q}}_n
=
g\!\left(
\widehat{\boldsymbol{\theta}}_n,
\mathbf{z}_n;
\boldsymbol{\kappa}
\right),
\]
where \(g(\cdot)\) is defined by Eq.~\eqref{eq:arrhenius_degradation}.
Because the battery measurements are sampled once per charge--discharge cycle, the degradation equation is integrated on the discrete cycle coordinate. Let \(\widehat{\dot{Q}}_{i,n}\) denote the degradation rate predicted for battery \(i\) at cycle \(n\). The capacity trajectory is computed using a first-order forward-Euler scheme:
\begin{equation}
\widehat{Q}_{i,n+1}
=
\widehat{Q}_{i,n}
+
\widehat{\dot{Q}}_{i,n}\Delta n,
\qquad
\Delta n = 1~\text{cycle},
\label{eq:capacity_euler}
\end{equation}
which is equivalently expressed as the cumulative sum
\begin{equation}
\widehat{Q}_{i,n}
=
Q_{i,0}
+
\sum_{j=0}^{n-1}
\widehat{\dot{Q}}_{i,j}\Delta n.
\label{eq:capacity_cumsum}
\end{equation}
Here, \(Q_{i,0}\) is the capacity measured at the first available cycle of battery \(i\) and provides the initial condition for integration. It is not predicted by the neural network. The numerical integration therefore uses one step per observed cycle and does not employ a higher-order Runge--Kutta method. The model-level output is the resulting predicted capacity or SOH trajectory, rather than the intermediate kinetic parameters themselves.

For the temporal-extrapolation experiment, the integration is initialized only once using \(Q_{i,0}\) at the beginning of each battery trajectory. The same forward-Euler recursion is continued from the observed training region into the held-out extrapolation region. The capacity is not reinitialized at the training--extrapolation boundary, and no measured capacity from the held-out region is introduced. Consequently, the first extrapolated capacity is calculated from the preceding predicted capacity, and the remaining extrapolated values are generated recursively from the predicted degradation rates.

Apart from the first measured capacity \(Q_{i,0}\), which provides the
initial condition for temporal integration, the experimentally observed capacity values are used only as supervised training targets and are not provided to the neural network as input features. The model minimizes the discrepancy between the predicted and observed capacity trajectories while enforcing the Arrhenius-based degradation structure. This formulation
guides the network toward physically consistent SOH evolution and improves its stability during long-horizon prediction.

\subsection{Baseline models, training procedure, and evaluation protocol}
\label{sec:baseline_training_evaluation}

To evaluate the contribution of the proposed physics-informed degradation formulation, PiDDM was compared with two neural-network baselines: an MLP and a baseline PINN. All three methods were evaluated using the same battery dataset, input-feature construction, normalization procedure, and evaluation metrics. Each neural-network component was implemented using fully connected layers with three hidden layers and 60 neurons per hidden layer.

\subsubsection{Data-driven MLP baseline}
\label{sec:mlp_baseline}

The MLP baseline represents a direct data-driven mapping from the
cycle-dependent charging features to the corresponding battery SOH. For battery
\(i\) at cycle \(n\), let
\(\mathbf{x}_{i,n}\in\mathbb{R}^{16}\) denote the normalized charging-feature
vector and let \(y_{i,n}\) denote the measured SOH. The MLP prediction is
written as

\begin{equation}
    \widehat{y}^{\,\mathrm{MLP}}_{i,n}
    =
    f_{\mathrm{MLP}}
    \left(
        \mathbf{x}_{i,n};
        \boldsymbol{\phi}_{\mathrm{MLP}}
    \right),
    \label{eq:mlp_prediction}
\end{equation}

where \(\boldsymbol{\phi}_{\mathrm{MLP}}\) denotes the trainable network
parameters. The MLP contains three fully connected hidden layers with
60 neurons per layer and a scalar output representing the predicted SOH.
Unlike PiDDM, the MLP does not contain an explicit degradation equation or
physics-informed temporal constraint.

The MLP is trained by minimizing the mean-squared error between the predicted
and measured SOH values,

\begin{equation}
    \mathcal{L}_{\mathrm{MLP}}
    =
    \frac{1}{N_{\mathrm{tr}}}
    \sum_{(i,n)\in\mathcal{D}_{\mathrm{tr}}}
    \left(
        \widehat{y}^{\,\mathrm{MLP}}_{i,n}
        -
        y_{i,n}
    \right)^2,
    \label{eq:mlp_loss}
\end{equation}

where \(\mathcal{D}_{\mathrm{tr}}\) denotes the training dataset and
\(N_{\mathrm{tr}}\) is the total number of training samples.


\subsubsection{Baseline PINN}
\label{sec:pinn_baseline}

The baseline PINN follows the battery degradation framework proposed
by Wang et al.~\cite{Wang_NC}. The model consists of two coupled
neural networks: a solution network, \(\mathcal{F}\), that maps the
charging features and cycle coordinate to SOH, and a dynamics network,
\(\mathcal{G}\), that learns the corresponding degradation law.

For battery \(i\) at cycle \(n\), let
\(\mathbf{x}_{i,n}\in\mathbb{R}^{16}\) denote the normalized charging
features and let \(\tau_{i,n}\) denote the normalized cycle coordinate.
The predicted SOH is given by

\begin{equation}
    \widehat{s}_{i,n}
    =
    \mathcal{F}
    \left(
        \tau_{i,n},
        \mathbf{x}_{i,n};
        \boldsymbol{\Phi}
    \right),
    \label{eq:pinn_solution}
\end{equation}

where \(\boldsymbol{\Phi}\) denotes the trainable parameters of the
solution network. The temporal and feature derivatives of
\(\widehat{s}\) are obtained through automatic differentiation and
provided to the dynamics network,

\begin{equation}
    \widehat{r}_{i,n}
    =
    \mathcal{G}
    \left(
        \tau_{i,n},
        \mathbf{x}_{i,n},
        \widehat{s}_{i,n},
        \frac{\partial \widehat{s}_{i,n}}{\partial \tau},
        \nabla_{\mathbf{x}}\widehat{s}_{i,n};
        \boldsymbol{\Theta}
    \right),
    \label{eq:pinn_dynamics}
\end{equation}

where \(\widehat{r}_{i,n}\) represents the learned SOH degradation
rate and \(\boldsymbol{\Theta}\) contains the trainable parameters of
the dynamics network. The physics-informed residual is then defined
as

\begin{equation}
    \mathcal{H}_{i,n}
    =
    \frac{\partial \widehat{s}_{i,n}}{\partial \tau}
    -
    \widehat{r}_{i,n}.
    \label{eq:pinn_residual}
\end{equation}

The two networks are trained jointly using a supervised data loss, a
degradation-residual loss, and a monotonicity penalty. The data loss
measures the discrepancy between the predicted and measured SOH,

\begin{equation}
    \mathcal{L}_{\mathrm{data}}
    =
    \sum_{(i,n)\in\mathcal{D}_{\mathrm{tr}}}
    \left|
        \widehat{s}_{i,n}-s_{i,n}
    \right|^2,
    \label{eq:pinn_data_loss}
\end{equation}

while the physics-informed loss constrains the predicted SOH
trajectory to be consistent with the learned degradation dynamics,

\begin{equation}
    \mathcal{L}_{\mathrm{PDE}}
    =
    \sum_{(i,n)\in\mathcal{D}_{\mathrm{tr}}}
    \left|
        \mathcal{H}_{i,n}
    \right|^2.
    \label{eq:pinn_pde_loss}
\end{equation}

Because battery capacity is expected to decrease with continued
cycling, a monotonicity penalty is applied to upward changes in the
predicted SOH,

\begin{equation}
    \mathcal{L}_{\mathrm{mono}}
    =
    \sum_i
    \sum_{n}
    \operatorname{ReLU}
    \left(
        \widehat{s}_{i,n+1}
        -
        \widehat{s}_{i,n}
    \right).
    \label{eq:pinn_monotonicity_loss}
\end{equation}

The complete training objective is

\begin{equation}
    \mathcal{L}_{\mathrm{PINN}}
    =
    \mathcal{L}_{\mathrm{data}}
    +
    \lambda_{\mathrm{PDE}}
    \mathcal{L}_{\mathrm{PDE}}
    +
    \lambda_{\mathrm{mono}}
    \mathcal{L}_{\mathrm{mono}},
    \label{eq:pinn_total_loss}
\end{equation}

where \(\lambda_{\mathrm{PDE}}\) and
\(\lambda_{\mathrm{mono}}\) control the contributions of the
degradation-residual and monotonicity terms, respectively. The loss
weights and network architectures follow the configuration reported
for the XJTU dataset by Wang et al.~\cite{Wang_NC}, with the complete
training settings summarized in
Section ~\ref{sec:training_procedure}

The baseline PINN and PiDDM incorporate degradation information in
fundamentally different ways. In the baseline PINN, the degradation
operator is represented by the flexible neural network
\(\mathcal{G}\) and inferred from the observed data. PiDDM instead
introduces the Arrhenius--power-law degradation relation in Eq.~\ref{eq:arrhenius_degradation}
as an explicit differentiable component of the forward model. The
comparison therefore evaluates whether prescribing physically
motivated degradation structure provides greater stability and
extrapolation capability than learning the degradation dynamics
through an unconstrained neural-network operator.

\subsubsection{PiDDM architecture}
\label{sec:piddm_architecture_details}

PiDDM receives the normalized 16-dimensional charging-feature vector as its input. The neural network maps the charging features to the parameters and dynamic variables required by the degradation relation in Eq.~(1). These outputs are then passed to the differentiable degradation module, which evaluates the capacity-fade rate. The predicted capacity trajectory is obtained through temporal integration of the resulting degradation rate.

Unlike the MLP, which directly maps the input features to SOH, PiDDM generates the predicted capacity trajectory through the degradation model. The Arrhenius--power-law relation therefore serves as a structural component of the PiDDM forward calculation and constrains the predicted trajectory toward physically consistent capacity fade.

%

\subsubsection{Network architectures}
\label{sec:network_architectures}

Table~\ref{tab:S2_architecture} summarizes the neural-network
architectures used for PiDDM, the MLP, and the baseline PINN. PiDDM
and the MLP use the same fully connected backbone, whereas the
baseline PINN follows the two-network architecture proposed by
Wang et al.~\cite{Wang_NC}.

\begin{table}[H]
\centering
\small
\caption{Neural-network architectures used for PiDDM, the MLP, and
the baseline PINN.}
\label{tab:S2_architecture}
\renewcommand{\arraystretch}{1.15}
\begin{tabularx}{\textwidth}{@{}P{4.2cm}P{2.2cm}L@{}}
\toprule
\textbf{Model component}
& \textbf{Input size}
& \textbf{Network architecture} \\
\midrule
PiDDM
& 16
& [16, 60, 60, 60, 5] \\

MLP
& 16
& [16, 60, 60, 60, 1] \\

PINN solution network, \(\mathcal{F}\)
& 17
& [17, 60, 60, 32, 32, 1] \\

PINN dynamics network, \(\mathcal{G}\)
& 35
& [35, 60, 60, 1] \\
\bottomrule
\end{tabularx}
\end{table}

The PINN solution network \(\mathcal{F}\) receives the 16 charging
features together with the cycle coordinate and predicts SOH. The
dynamics network \(\mathcal{G}\) represents the learned degradation
operator and receives the predicted SOH, the cycle-dependent
features, and the corresponding first-order derivatives obtained
through automatic differentiation. Sine activation functions are
applied within the hidden layers of both PINN components, following
the original implementation~\cite{Wang_NC}. The dynamics network is
used to constrain model training, whereas SOH estimation during
inference is performed using the solution network.

The architectures were not forced to have identical layer structures.
Instead, PiDDM and the MLP share the same data-driven backbone to
isolate the effect of the differentiable degradation model, while the
baseline PINN retains the architecture of the published method. The
comparison is conducted using the same input information, data
partitions, preprocessing procedure, and evaluation metrics.

\subsubsection{Training procedure}
\label{sec:training_procedure}

All models were optimized using the Adam optimizer. A learning-rate
warmup schedule was applied during the first 30 epochs, followed by
cosine decay to the final learning rate. PiDDM and the MLP were
trained for 500 epochs. The solution and degradation-dynamics networks of the PINN were
optimized jointly.

As described in Section~\ref{sec:pinn_baseline}, the baseline PINN
objective combines the supervised SOH loss, the degradation-residual
loss, and the monotonicity loss,

\begin{equation}
    \mathcal{L}_{\mathrm{PINN}}
    =
    \mathcal{L}_{\mathrm{data}}
    +
    \lambda_{\mathrm{PDE}}
    \mathcal{L}_{\mathrm{PDE}}
    +
    \lambda_{\mathrm{mono}}
    \mathcal{L}_{\mathrm{mono}}.
    \label{eq:pinn_training_objective}
\end{equation}

For the dataset considered in this work, the loss weights were
set to \(\lambda_{\mathrm{PDE}}=0.7\) and
\(\lambda_{\mathrm{mono}}=20\), following Wang et
al.~\cite{Wang_NC}. The complete training settings are summarized in
Table~\ref{tab:S1_training}.

\begin{table}[H]
\centering
\small
\caption{Training hyperparameters for PiDDM, the MLP, and the
baseline PINN.}
\label{tab:S1_training}
\renewcommand{\arraystretch}{1.15}
\begin{tabularx}{\textwidth}{@{}P{4.0cm}P{4.0cm}P{4.0cm}@{}}
\toprule
\textbf{Hyperparameter}
& \textbf{PiDDM \& MLP}
& \textbf{Baseline PINN} \\
\midrule
Epochs
& 500
& 200 \\

Batch size
& 10000
& 256 \\

Base learning rate
& 0.001
& 0.01 \\

Final learning rate
& 0.0002
& 0.0002 \\

Warmup epochs
& 30
& 30 \\

Warmup learning rate
& 0.002
& 0.002 \\

Normalization method
& Min--max
& Min--max \\

PDE-loss weight, \(\lambda_{\mathrm{PDE}}\)
& Not applicable
& 0.7 \\

Monotonicity-loss weight, \(\lambda_{\mathrm{mono}}\)
& Not applicable
& 20 \\
\bottomrule
\end{tabularx}
\end{table}

All input features and SOH values were normalized using training-set
statistics. For a variable \(x\), min--max normalization is expressed
generally as

\begin{equation}
    \widetilde{x}
    =
    2 \cdot
    \frac{x-x_{\min}^{\mathrm{tr}}}
         {x_{\max}^{\mathrm{tr}}-x_{\min}^{\mathrm{tr}}}-1,
    \label{eq:minmax_normalization}
\end{equation}

where \(x_{\min}^{\mathrm{tr}}\) and
\(x_{\max}^{\mathrm{tr}}\) are calculated exclusively from the
training data, and \([a,b]\) denotes the prescribed normalization
interval. The resulting transformation was applied unchanged to the
validation and testing data to prevent information leakage.

\subsubsection{Evaluation protocol}
\label{sec:evaluation_protocol}

The models were trained and evaluated using the public NCM battery dataset reported by Wang et al.~\cite{Wang_NC}. The dataset contains 55 batteries cycled to failure under six operating protocols: constant-current 2C and 3C protocols, randomized-discharge R2.5 and R3 protocols, a random-walk (RW) protocol, and a geosynchronous Earth orbit satellite-mission (Sim) protocol. The 3C protocol contains 15 batteries, numbered 1-15; batteries 4 and 8 were reserved for testing, while the remaining 13 batteries were used for training. Each of the other five protocols contains eight batteries; batteries 1, 2, 3, 5, 6, and 7 were used for training, while batteries 4 and 8 were reserved for testing. This battery-level partition resulted in 43 training batteries and 12 independent test batteries, accounting for all 55 batteries in the dataset. PiDDM, the MLP, and the baseline PINN were evaluated using this identical partition in every experiment.

Two evaluation settings were considered. In the standard prediction setting, the models were trained using the complete trajectories assigned to the training dataset and evaluated on batteries excluded from model training. This experiment evaluates the ability of each method to estimate SOH across the six operating protocols.

In the temporal-extrapolation setting, only the initial \(90\%\) of each degradation trajectory was included during training. The final \(10\%\) was excluded from optimization and treated as the unseen extrapolation region. No SOH labels from this held-out region were used to optimize the models.

Model performance was quantified using mean absolute percentage error (MAPE), root mean squared error (RMSE), mean absolute error (MAE), and mean squared error (MSE). For \(N_{\mathrm{te}}\) evaluation samples, these metrics are defined as

\begin{align}
    \mathrm{MAPE}
    &=
    \frac{1}{N_{\mathrm{te}}}
    \sum_{j=1}^{N_{\mathrm{te}}}
    \left|
        \frac{\widehat{y}_{j}-y_{j}}{y_{j}}
    \right|,
    \label{eq:mape_metric}
    \\[3pt]
    \mathrm{RMSE}
    &=
    \sqrt{
        \frac{1}{N_{\mathrm{te}}}
        \sum_{j=1}^{N_{\mathrm{te}}}
        \left(
            \widehat{y}_{j}-y_{j}
        \right)^2
    },
    \label{eq:rmse_metric}
    \\[3pt]
    \mathrm{MAE}
    &=
    \frac{1}{N_{\mathrm{te}}}
    \sum_{j=1}^{N_{\mathrm{te}}}
    \left|
        \widehat{y}_{j}-y_{j}
    \right|,
    \label{eq:mae_metric}
    \\[3pt]
    \mathrm{MSE}
    &=
    \frac{1}{N_{\mathrm{te}}}
    \sum_{j=1}^{N_{\mathrm{te}}}
    \left(
        \widehat{y}_{j}-y_{j}
    \right)^2,
    \label{eq:mse_metric}
\end{align}

where \(y_j\) and \(\widehat{y}_j\) denote the measured and predicted SOH,
respectively. Lower values indicate improved prediction
accuracy.

\section{Results}

The predictive performance of the PiDDM framework is evaluated through two sets of experiments. First, we compare PiDDM with baseline deep-learning models to assess its accuracy across multiple cycling protocols. Second, we examine its extrapolation capability by training the models only on the early portion of the battery life and evaluating their performance on the remaining unseen cycles. Through comparison with purely data-driven MLP and baseline PINN models, these experiments highlight the role of physics-based regularization in improving both predictive accuracy and physical consistency.

\subsection{PiDDM performance}

To evaluate the predictive accuracy of PiDDM, we compare it with a purely data-driven MLP and a baseline PINN. Table~\ref{tab:comparison} summarizes the prediction errors in terms of MAPE, RMSE, MAE, and MSE across six battery aging protocols.

\begin{table}[!ht]
\centering
\caption{Comparison of prediction error metrics across models and datasets. Performance of MLP, baseline PINN, and the proposed PiDDM is evaluated on six battery testing protocols (2C, 3C, R2.5, R3, RW, and Satellite). The lowest value for each metric and protocol is highlighted in bold.}
\label{tab:comparison}
\includegraphics[width=\textwidth]{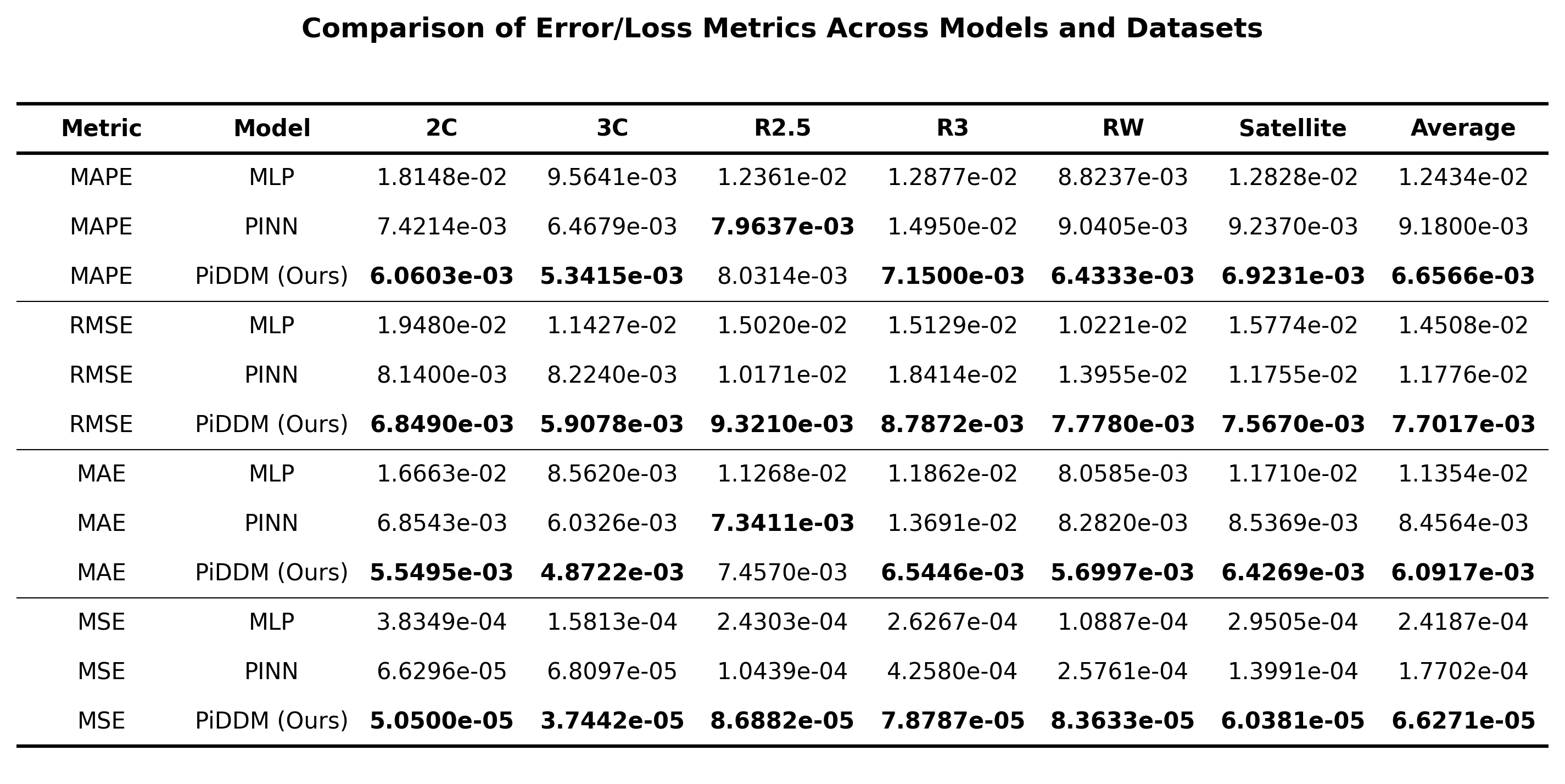}
\end{table}

Overall, PiDDM achieves the lowest average error across all four metrics, indicating improved generalization across diverse operating conditions. In particular, the average MSE of PiDDM is $6.6271 \times 10^{-5}$, compared with $1.7702 \times 10^{-4}$ for the baseline PINN and $2.4187 \times 10^{-4}$ for the MLP. The improvement is especially evident for the highly dynamic RW and Satellite protocols, where the auto-regressive structure of PiDDM better captures the temporal evolution of capacity degradation.

\begin{figure}[!ht]
\centering
\includegraphics[width=\textwidth]{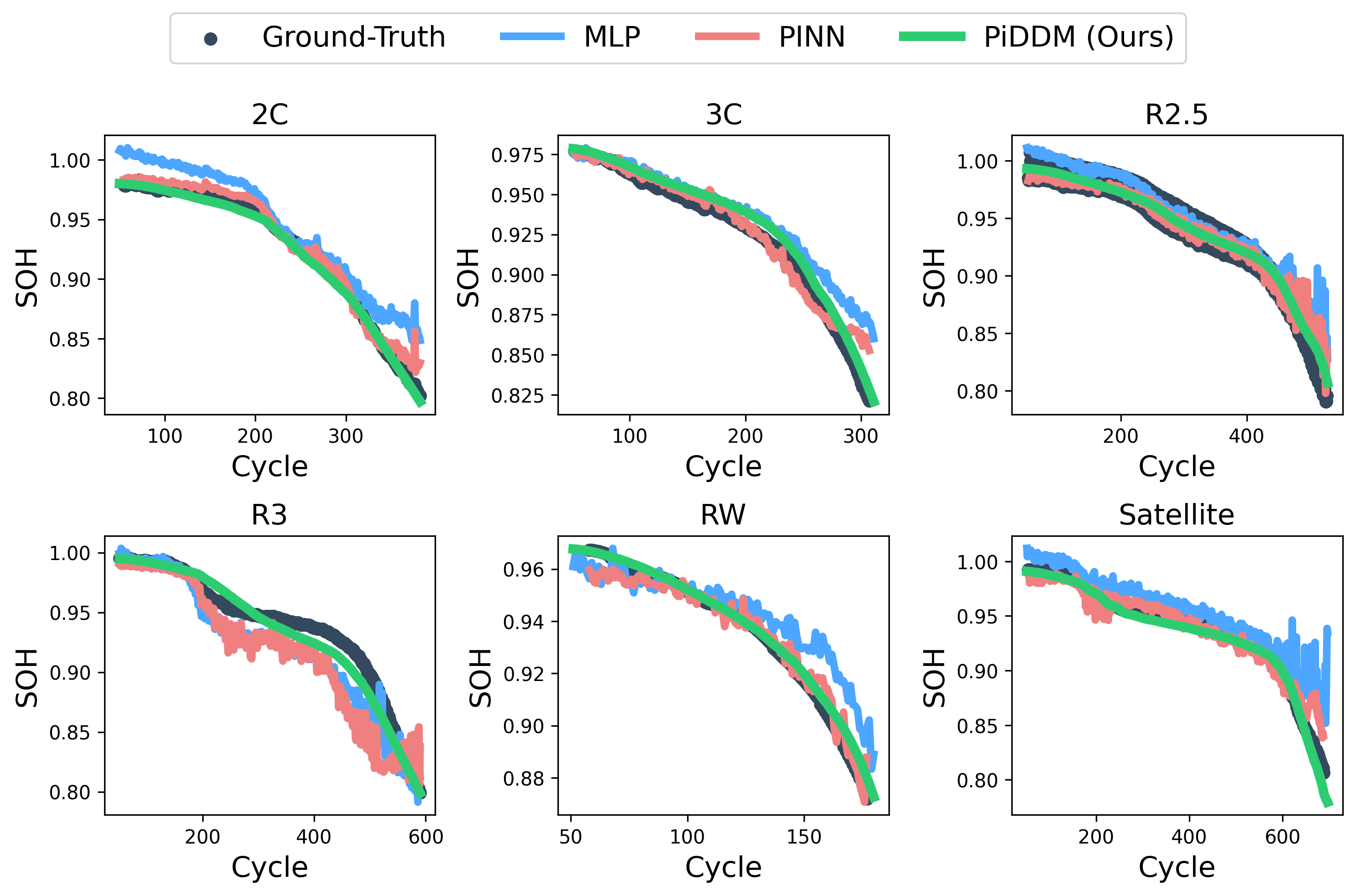}
\caption{\textbf{Predictive performance of PiDDM across diverse protocols.} SOH predictions from PiDDM (green) are compared with those of MLP (blue) and PINN (red) across six cycling protocols: constant current (2C, 3C), randomized discharge (R2.5, R3), random walk (RW), and satellite missions.}
\label{fig:comparison_SOH}
\end{figure}

Although PiDDM performs best on most protocols and metrics, a small deviation is observed for the R2.5 protocol. In this case, the baseline PINN achieves slightly lower MAPE ($7.9637 \times 10^{-3}$) and MAE ($7.3411 \times 10^{-3}$) than PiDDM ($8.0314 \times 10^{-3}$ and $7.4570 \times 10^{-3}$, respectively). As shown in Figure~\ref{fig:comparison_SOH}, the ground-truth SOH trajectory for R2.5 contains localized upward fluctuations, which are likely associated with measurement noise or temporary regeneration artifacts. Because PiDDM is regularized by degradation physics that enforces non-increasing SOH trajectories, it does not track these local upward variations as closely as the baseline PINN. As a result, its absolute mean error is slightly higher for this protocol.

Despite this behavior, PiDDM still yields lower RMSE ($9.3210 \times 10^{-3}$) and MSE ($8.6882 \times 10^{-5}$) for the R2.5 dataset. Because RMSE and MSE penalize larger deviations more strongly than MAE and MAPE, these results indicate that PiDDM remains less sensitive to large prediction errors and outliers, even when the absolute mean errors are comparable. Taken together, the results in Table~\ref{tab:comparison} show that integrating physics-based regularization improves the robustness of SOH prediction across diverse cycling conditions.

Figure~\ref{fig:comparison_SOH} provides a qualitative comparison of the predicted SOH trajectories. The purely data-driven MLP exhibits noticeable variance and does not consistently capture the nonlinear degradation trends, particularly during the later stages of cycling. The baseline PINN produces smoother trajectories, but it still shows visible deviations from the ground truth and does not always capture the onset of the end-of-life degradation knee accurately. In contrast, PiDDM more closely follows the true trajectories across all six protocols, capturing both the gradual degradation phase and the accelerated end-of-life decline with improved stability.

\subsection{Extrapolation performance}

\begin{figure}[!ht]
\centering
\includegraphics[width=\textwidth]{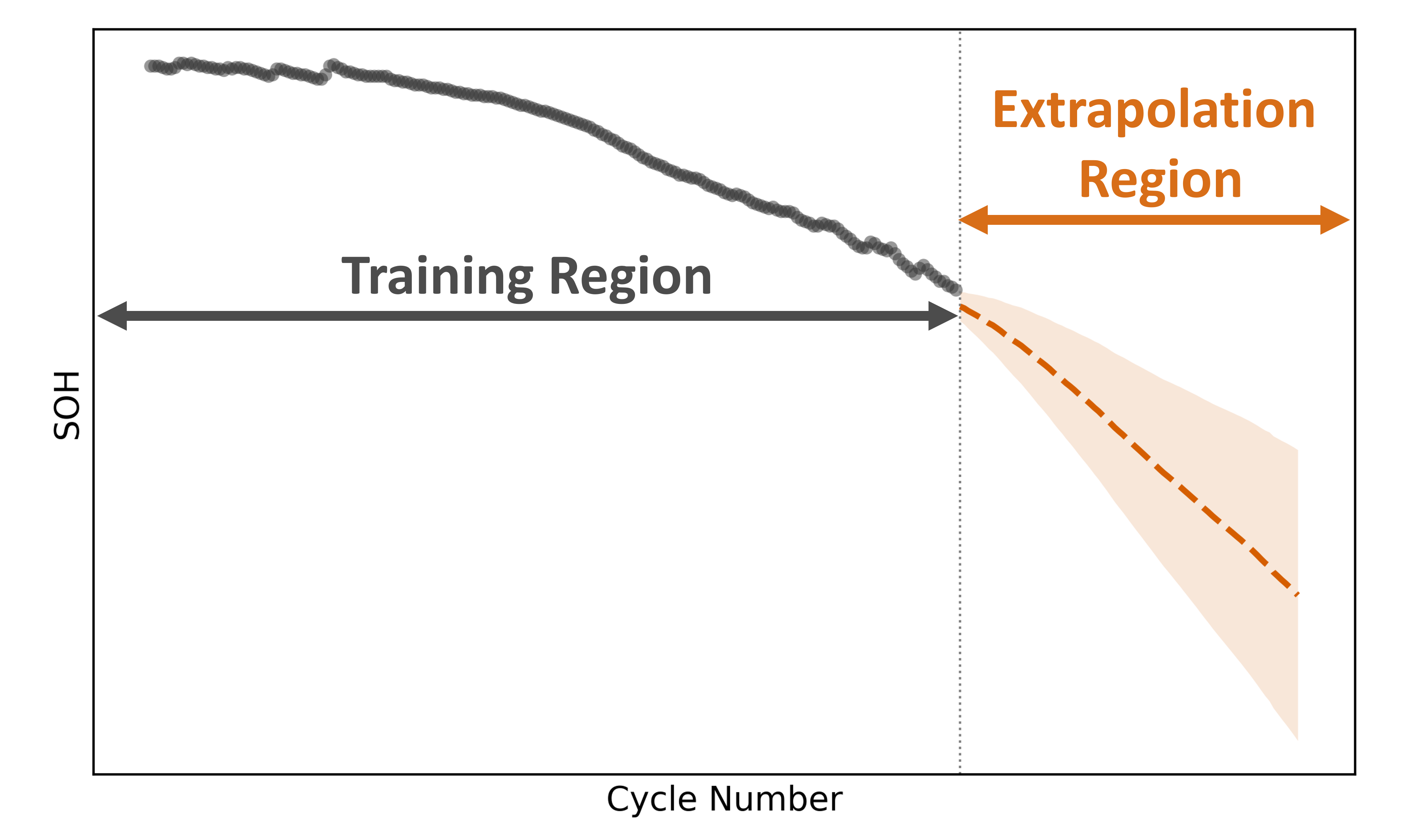}
\caption{\textbf{Extrapolation evaluation protocol for SOH prediction.} Observed degradation data (gray points) are used for training, while the extrapolation region corresponds to the unseen portion of the SOH trajectory that must be predicted by the model.}
\label{fig:extrap_setup}
\end{figure}

To assess the extrapolation capability of PiDDM, we evaluate the models under an extrapolation setting in which only the early portion of the degradation trajectory is available during training. Figure~\ref{fig:extrap_setup} illustrates the evaluation setup. For each battery in the training dataset, the SOH trajectory is divided into an initial training region and a subsequent extrapolation region. Only the data in the training region are used for model fitting. After training, the models are evaluated using the independent testing dataset, which contains complete trajectories spanning both the corresponding early-cycle region and the final extrapolation region. The final portion of each trajectory is therefore temporally outside the range used for model training. This setup tests whether the learned models can extend the observed degradation behavior and produce physically plausible predictions beyond the temporal range represented during training.

Under the extrapolation setup described above, the models are trained using only the initial 90\% of each available training trajectory and are required to predict beyond this temporal range during testing. Table~\ref{tab:extrap_compare} summarizes the prediction errors under this extrapolation condition. PiDDM achieves the lowest average error across all four metrics, with an average MSE of \(1.1067\times10^{-4}\), compared with \(2.4381\times10^{-4}\) for the baseline PINN and \(8.2631\times10^{-4}\) for the MLP. PiDDM also achieves the lowest errors for five of the six protocols: 2C, 3C, R2.5, RW, and Satellite. The R3 protocol is the exception, for which the MLP attains the lowest values across all four reported metrics, including an MSE of \(1.5786\times10^{-4}\), compared with \(3.4648\times10^{-4}\) for PiDDM. Thus, the MLP provides greater pointwise prediction accuracy than PiDDM for R3, although PiDDM performs better on average and across the other five protocols.

\begin{table}[!ht]
\centering
\caption{Comparison of prediction error metrics under extrapolation. Models are trained on the initial 90\% of each battery's cycle life and evaluated on the remaining unseen 10\%. The lowest value for each metric and protocol is highlighted in bold.}
\label{tab:extrap_compare}
\includegraphics[width=\textwidth]{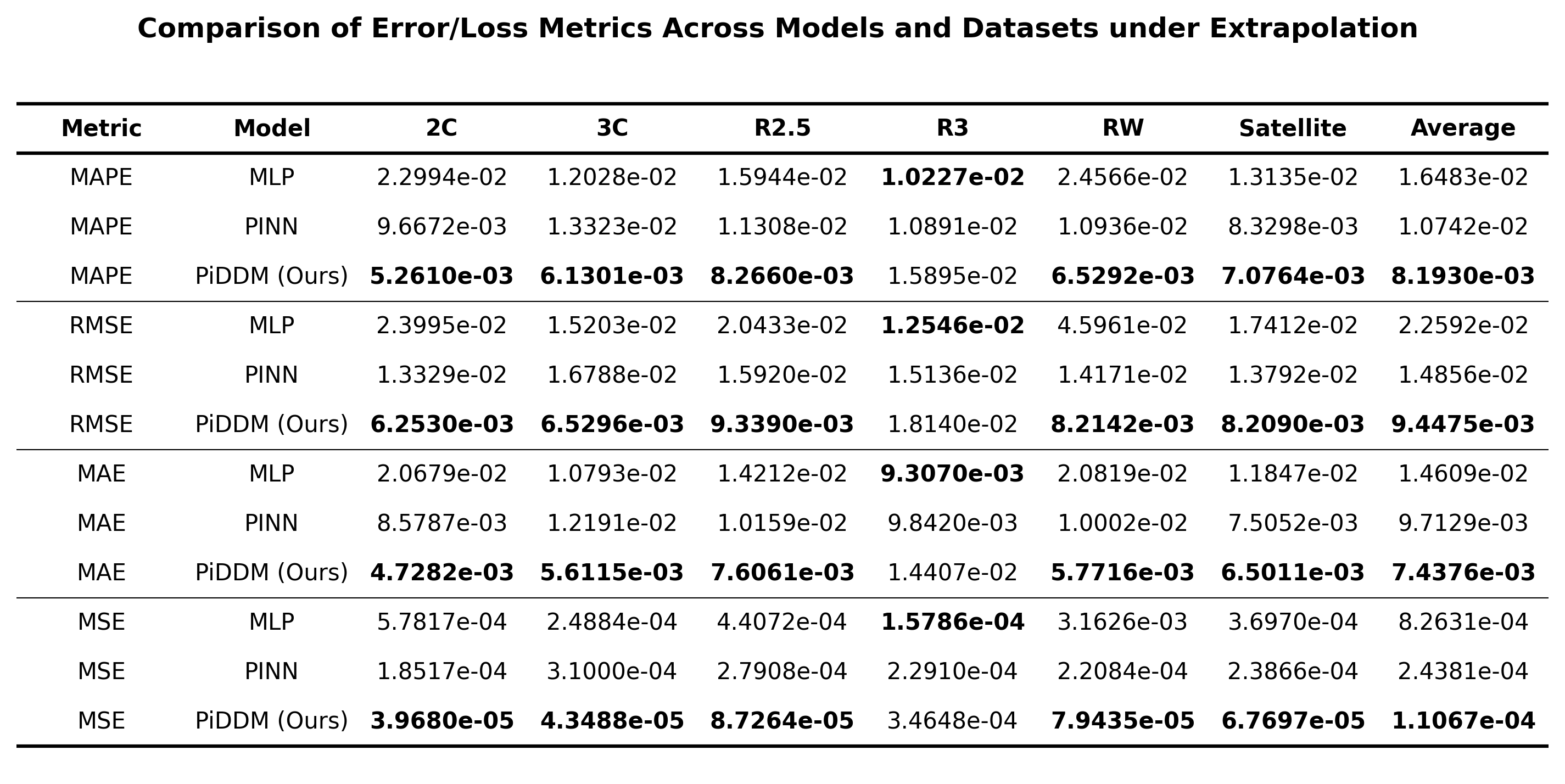}
\end{table}

\begin{figure}[!ht]
\centering
\includegraphics[width=\textwidth]{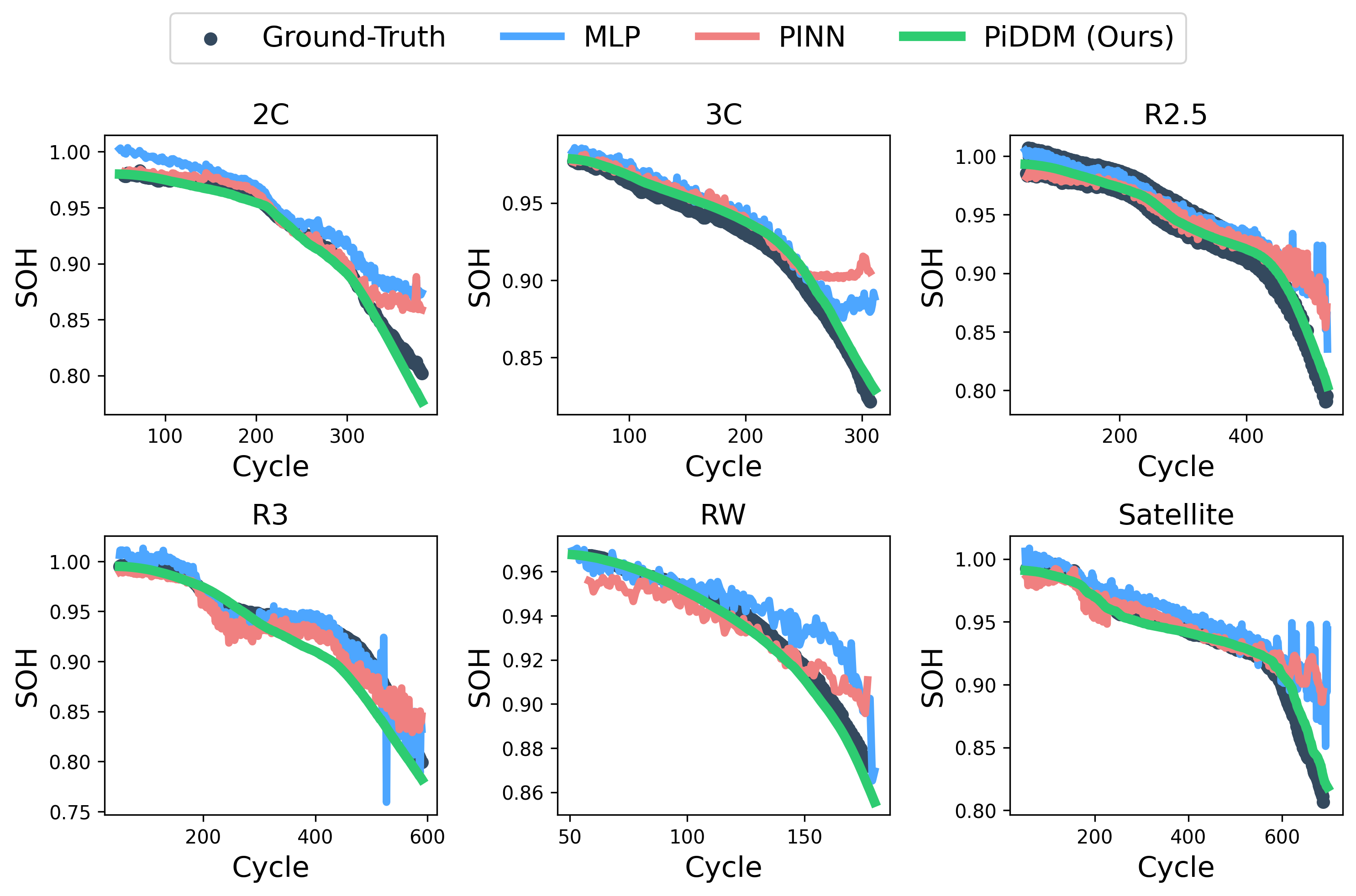}
\caption{\textbf{Temporal extrapolation performance across diverse cycling protocols.} SOH predictions from MLP, PINN, and PiDDM are compared across six protocols: varying C-rates (2C and 3C), resistance-based aging conditions (R2.5 and R3), and dynamic load profiles (RW and Satellite). Models are trained on the initial 90\% of the cycle life and evaluated on the remaining unseen 10\%.}
\label{fig:extrap_SOH}
\end{figure}

To provide additional qualitative insight into the model predictions, Figure~\ref{fig:extrap_SOH} compares the extrapolated SOH trajectories produced by the three models. In the following discussion, trajectory instability refers specifically to visually apparent cycle-to-cycle spikes, high-frequency noise, and local upward reversals in the predicted SOH trajectory; it does not refer to numerical divergence during training and is not introduced as an additional quantitative metric. The ground-truth R3 trajectory exhibits an accelerated capacity drop near the end of life. The MLP remains sufficiently close to the measured R3 values to achieve the lowest aggregate prediction error, but its predicted trajectory also contains pronounced noise and a large downward spike followed by an upward reversal near the end of the extrapolation region. These observations are not contradictory: the error metrics quantify the average discrepancy between the predicted and measured SOH values, whereas the trajectory plot also reveals the smoothness and directional consistency of the cycle-to-cycle predictions.

In comparison, PiDDM produces a smoother and more nearly monotonic R3 prediction, although its pointwise error is higher for this protocol. The Arrhenius-based degradation module constrains the predicted capacity-fade rate through the assumed temperature-dependent degradation kinetics, thereby discouraging abrupt cycle-to-cycle changes and pronounced upward reversals. The R3 result should therefore be interpreted as a distinction between two aspects of prediction quality rather than as an unqualified improvement by either model: the MLP more accurately reproduces the measured R3 values according to the four reported error metrics, whereas PiDDM more clearly preserves the expected long-term direction and smoothness of cumulative capacity degradation.

Similar qualitative behavior can be observed across the other protocols in Figure~\ref{fig:extrap_SOH}. The MLP produces visible noise or isolated spikes in several late-stage predictions, while the baseline PINN exhibits pronounced upward spikes for protocols such as 3C, R2.5, and Satellite. PiDDM largely suppresses these artifacts and produces smoother extrapolated trajectories while capturing the accelerated degradation knee. Because measured SOH can contain experimental noise and temporary apparent capacity recovery, an isolated increase should not automatically be interpreted as physical capacity regeneration. Nevertheless, the reduced occurrence of large spikes and short-term reversals indicates that PiDDM follows the underlying cumulative degradation trend more consistently. Overall, the quantitative results demonstrate that PiDDM achieves the best average extrapolation accuracy and the lowest errors for five of the six protocols, while the trajectory plots qualitatively show smoother and more physically consistent long-horizon predictions. The exception is R3, for which the MLP remains the most accurate model according to all four reported error metrics.


\section{Conclusion and Discussion}

In this work, we demonstrated a physics-informed differentiable degradation modeling  (PiDDM) for lithium-ion battery state-of-health (SOH) prediction under diverse cycling conditions. The framework combines data-driven learning with an empirical degradation law that accounts for capacity fade associated with solid electrolyte interphase growth and loss of lithium inventory. By incorporating this degradation relation into the training objective, PiDDM improves the physical consistency of the predicted SOH trajectories while maintaining strong predictive performance across varied operating protocols.

The model was evaluated on the public battery dataset, which contains 55 batteries cycled under six distinct charge and discharge protocols. Compared with a purely data-driven MLP and a baseline PINN, PiDDM achieved lower average prediction error across standard prediction tasks. The improvement was particularly evident under dynamic operating conditions, such as the random-walk and satellite protocols, where the combined effects of temporal dependence and protocol variability make accurate prediction more challenging.

PiDDM also showed improved extrapolation performance when the models were trained only on the initial 90\% of each battery's cycle life and evaluated on the remaining unseen 10\%. Under this setting, the proposed framework produced smoother and more physically consistent degradation trajectories than the baseline models, while more accurately capturing the accelerated end-of-life degradation knee. In contrast, the MLP exhibited unstable oscillatory behavior during long-horizon prediction, and the baseline PINN occasionally produced non-physical upward fluctuations in SOH. These results indicate that incorporating Arrhenius-based degradation regularization can improve both the stability and physical plausibility of long-horizon battery health prediction.

At the same time, the results also highlight an important tradeoff. Because PiDDM is regularized toward physically consistent degradation behavior, it does not always follow localized upward fluctuations in the measured SOH trajectory as closely as less constrained models. As observed for the R2.5 protocol, this can lead to a slightly higher absolute error in cases where the experimental data contain measurement noise or temporary regeneration-like artifacts. However, this behavior is beneficial for practical battery health monitoring, where robustness and physical consistency are often more important than fitting local fluctuations that may not reflect true degradation.

Overall, this study shows that embedding empirical degradation physics into neural network training provides an effective route toward more reliable battery SOH prediction across both standard and extrapolative settings. Future work will focus on extending the framework to incorporate additional operating variables and degradation indicators, such as time-varying temperature and internal resistance, and on evaluating its performance across a broader range of battery chemistries and usage conditions.


\section*{Data and code availability}
Data and code are available upon request.

\section*{Acknowledgment}
G.X. would like to acknowledge the funds from the National Science Foundation (NSF) grant (No. 2533718) in supporting this study.

\section*{Author Contributions}
Z.C., G.X., and T.L. contributed to the ideation and design of the research; 
Z.C. G.X., and T.L. performed the research; Z.C. wrote the manuscript; 
R.J., S.S., G.X., and T.L. contributed to manuscript editing.

\section*{Competing interests}
The authors declare no competing interests.

\noindent\textbf{Corresponding authors:} Tengfei Luo (\url{tluo@nd.edu}).
\bibliographystyle{elsarticle-num}
\bibliography{ref}
\end{document}